\documentclass{article}

\usepackage[preprint]{neurips_2025}

\usepackage[utf8]{inputenc} 
\usepackage[T1]{fontenc}    
\usepackage{hyperref}       
\usepackage{url}            
\usepackage{booktabs}       
\usepackage{amsfonts}       
\usepackage{nicefrac}       
\usepackage{microtype}      
\usepackage{xcolor}         
\usepackage{amsmath}

\usepackage{graphicx}
\usepackage{subcaption} 

\title{Parallel qMRI Reconstruction from 4x Accelerated Acquisitions}

\author{%
  Mingi Kang \\
  Bowdoin College\\
  Brunswick, ME 04011 \\
  \texttt{mkang2@bowdoin.edu} \\
}

\begin{document}

\maketitle

\begin{abstract}
Magnetic Resonance Imaging (MRI) acquisitions require extensive scan times, limiting patient throughput and increasing susceptibility to motion artifacts. Accelerated parallel MRI techniques reduce acquisition time by undersampling k-space data, but require robust reconstruction methods to recover high-quality images. Traditional approaches like SENSE require both undersampled k-space data and pre-computed coil sensitivity maps. We propose an end-to-end deep learning framework that jointly estimates coil sensitivity maps and reconstructs images from only undersampled k-space measurements at 4x acceleration. Our two-module architecture consists of a Coil Sensitivity Map (CSM) estimation module and a U-Net-based MRI reconstruction module. We evaluate our method on multi-coil brain MRI data from 10 subjects with 8 echoes each, using 2x SENSE reconstructions as ground truth. Our approach produces visually smoother reconstructions compared to conventional SENSE output, achieving comparable visual quality despite lower PSNR/SSIM metrics. We identify key challenges including spatial misalignment between different acceleration factors and propose future directions for improved reconstruction quality. 
\end{abstract}


\begin{center}
  \url{https://github.com/mingikang31/MRI-Reconstruction}
\end{center}

\section{Introduction}
Magnetic Resonance Imaging (MRI) is a fundamental diagnostic tool in modern medicine, offering non-invasive visualization of internal anatomical structures. However, full MRI acquisitions require extensive scan times, often lasting 30-60 minutes per patient. These prolonged acquisition times limit patient throughput, increase healthcare costs, and make patients more susceptible to motion artifacts that degrade image quality. Accelerating MRI acquisitions while maintaining diagnostic image quality remains a critical challenge in medical imaging. 

Parallel MRI techniques address this challenge by leveraging multiple receiver coils to undersample k-space data, thereby reducing acquisition time \cite{pruessmann1999sense}. In 4x accelerated acquisitions, only every fourth point of k-space is measured, theoretically reducing scan time by a factor of four. However, this undersampling introduces aliasing artifacts that must be corrected during reconstruction. Traditional reconstruction methods such as SENSE (SENSitivity Encoding) \cite{pruessmann1999sense} and GRAPPA (GeneRalized Autocalibrating Partial Parallel Acquisition) \cite{griswold2002grappa} have been widely adopted, but these approaches have limitations. SENSE requires accurate pre-computed coil sensitivity maps in addition to the undersampled k-space data, necessitating additional calibration scans or separate estimation procedures. 

Recent advances in deep learning have demonstrated remarkable success in various image reconstruction tasks \cite{hammernik2018variational, zbontar2018fastmri}. U-Net architectures, originally developed for biomedical image segmentation \cite{ronneberger2015unet}, have shown particular promise for MRI reconstruction due to their ability to capture both local and global image features through their encoder-decoder structure with skip connections. Several works have explored deep learning for MRI reconstruction, including data-driven approaches that learn mappings from undersampled to fully-sampled images \cite{zbontar2018fastmri}. 

In this work, we propose an end-to-end deep learning framework for parallel MRI reconstruction from 4x accelerated acquisitions. Our key contribution is a two-module architecture that jointly performs coil sensitivity map estimation and image reconstruction, requiring only undersampled k-space measurements as input. This work extends SPICER (Self-supervised learning for MRI with automatic coil sensitivity estimation and reconstruction) \cite{hu2024spicer}, which employs a deep unfolding network with multiple sequential U-Net architectures. Unlike SPICER's complex multi-stage unfolding, our approach uses a simplified two-module design, trading some reconstruction accuracy for improved computational efficiency and easier optimization. Additionally, unlike SENSE, our two-module approach eliminates the need for separate sensitivity map acquisition or pre-computation. We evaluate our method on clinical brain MRI data and analyze the performance differences between simulated and real 4x accelerated acquisitions, identifying important practical challenges for deployment. 

\section{Related Work}
\subsection{Traditional Parallel MRI Reconstruction}
Parallel imaging techniques utilize spatial encoding from multiple receiver coils to accelerate MRI acquisitions. SENSE \cite{pruessmann1999sense} reconstructs images by solving a linear inverse problem that incorporates coil sensitivity information, effectively "unfolding" aliased pixels. GRAPPA \cite{griswold2002grappa} operates in k-space, using acquired data to estimate missing k-space lines through linear combinations. While effective, these methods can produce residual aliasing artifacts, particularly at higher acceleration factors, and require accurate sensitivity map estimation or sufficient calibration data. 

\subsection{Deep Learning for MRI Reconstruction}
Recent years have witnessed growing interest in applying deep learning to MRI reconstruction. Early approaches treated reconstruction as an image-to-image translation problem, using convolutional neural networks to map zero-filled reconstructions to artifact-free images \cite{zbontar2018fastmri}. The U-Net architecture \cite{ronneberger2015unet} has become particularly popular due to its encoder-decoder structure with skip connections, which preserves fine details while capturing global context. Attention U-Net \cite{oktay2018attention} further enhanced this architecture by incorporating attention gate mechanisms to focus on relevant image regions. 

\subsection{Learning-Based Parallel Imaging}
Several works have explored combining deep learning with parallel imaging principles. Variational networks \cite{hammernik2018variational} unroll iterative reconstruction algorithms into trainable networks. SPICER \cite{hu2024spicer} proposed self-supervised learning for MRI with automatic coil sensitivity estimation, demonstrating that joint optimization of sensitivity maps and image reconstruction can improve results. Our work builds on these ideas but focuses specifically on practical deployment with real 4x accelerated clinical data, analyzing the challenges that arise when transitioning from simulated to real undersampled acquisitions. 

\section{Methodology}
\subsection{Problem Formulation}
In parallel MRI, the forward model relating the image $x \in \mathbb{C}^{H \times W}$ to the measured k-space data $y \in \mathbb{C}^{C \times H \times W}$ for $C$ coils is: 

\begin{equation}
    y = M\mathcal{F}Sx + n
\end{equation}

where $\mathcal{F}$ is the 2D Fourier transform operator, $S \in \mathbb{C}^{C \times H \times W}$ represents the coil sensitivity maps, $M \in \{0, 1\}^{H \times W}$ is the undersampling mask, and $n$ is measurement noise. For 4x acceleration, $M$ uniformly undersamples the phase-encoding direction by a factor of 4, while retaining a fully-sampled Auto-Calibration Signal (ACS) region of size $48 \times 48$ at the k-space center for improved reconstruction stability.

Our goal is to reconstruct a high-quality image $\hat{x}$ from undersampled measurements $y$ without requiring pre-computed sensitivity maps $S$. We achieve this through a two-stage approach: first estimating sensitivity maps, then reconstructing the final image using both the estimated maps and the undersampled data. 

\begin{table}[tb]
\centering
\caption{Quantitative reconstruction results on test set.}
\label{tab:metric_eval}
\begin{tabular}{lccc}
\toprule
Method & Input Type & PSNR (dB) & SSIM \\
\midrule
SENSE & Real 4x      & $35.97  \pm 1.46$ & $0.937 \pm 0.023$ \\
Ours & Real 4x       & $33.24 \pm 2.11$ & $0.879 \pm 0.051$ \\
Ours & Simulated 4x  & $35.06 \pm 2.11$ & $0.900 \pm 0.045$ \\
\bottomrule
\end{tabular}

\vspace{2pt}
{\small All values presented as mean $\pm$ standard deviation}
\end{table}

\subsection{Network Architecture}
Our framework consists of two modules that operate sequentially, with a total parameter count of approximately 31.5M for a 4-layer pooling architecture starting with 64 channels.

\subsubsection{Coil Sensitivity Map Estimation Module}
This module takes the undersampled k-space data $y$ as input and estimates coil sensitivity maps $\hat{S}$. The module uses a U-Net architecture with the following structure:
\begin{itemize}
    \item Input: Multi-coil k-space data ($64 \text{ coils } \times 234 \times 176$)
    \item Encoder: 4 downsampling blocks with $[64, 128, 256, 512]$ channels
    \item Decoder: 4 upsampling blocks with skip connections
    \item Output: Estimated sensitivity maps $\hat{S}$ ($64 \text{ coils } \times 234 \times 176$)
\end{itemize}

\subsubsection{MRI Reconstruction Module}
This module performs the final image reconstruction by first computing a composite k-space:

\begin{equation}
    k^+ = (k \odot M) + \lambda((1 - M) \odot \mathcal{F}(S_m R_m(S_m^+ \mathcal{F}^{-1}k)))
\end{equation}

where $\odot$ denotes element-wise multiplication, $S_m$ represents the estimated sensitivity maps, $R_m$ is the root-sum-of-squares combination operator, and $\lambda$ is a weighting parameter that controls the contribution of the estimated missing k-space data. The composite k-space $k^+$ combines the measured k-space values (where $M=1$) with estimated values for unsampled locations (where $M=0$).

By applying the inverse Fourier transform $\mathcal{F}^{-1}$ to the composite k-space, we obtain an initial image reconstruction. This initial reconstruction is then refined using a U-Net denoiser with:

\begin{itemize}
    \item Input: Initial reconstruction ($234 \times 176$)
    \item Encoder-Decoder structure with $[32, 64, 128, 256]$ channels 
    \item Skip connections between encoder and decoder 
    \item Output: Final reconstructed image $\hat{x}$ ($234 \times 176$)
\end{itemize}

The parameter $\lambda$ was set as a learnable parameter in our experiments.

\subsection{Loss Function}
We train the network using a composite loss that operates in both image and k-space domains: 

\begin{equation}
    L_{\text{total}} = 1.5 \cdot L_{\text{img}} + 0.5 \cdot L_{\text{ksp}}
\end{equation}

where $L_{\text{img}} = \text{MSE}(\hat{x}, x_{\text{gt}})$ measures image-domain reconstruction error and $L_{\text{ksp}} = \text{MSE}(\mathcal{F}S\hat{x}, y_{\text{gt}})$ ensures k-space consistency.

The weights (1.5 and 0.5) were chosen to emphasize image-domain quality while maintaining k-space fidelity. Ground truth images $x_{\text{gt}}$ are obtained from 2x SENSE reconstructions, which serve as our reference standard. 
\subsection{Training Details}
\subsubsection{Data Generation}
We created simulated 4x accelerated data by applying a 4x undersampling mask to the SENSE-reconstructed 2x accelerated k-space measurements. This allows direct comparison with real 4x acquired data while maintaining perfect alignment with the ground truth. Importantly, the original 2x accelerated data contains an ACS region of $32 \times 32$, while the 4x accelerated data has a larger ACS region of $48 \times 48$. Due to this difference in acquisition patterns, we reserved the true 4x accelerated measurements as test data, while training data was generated by applying 4x undersampling masks to the reconstructed 2x k-space data.

\subsubsection{Optimization}
We trained the network end-to-end using the Adam optimizer with $\beta_1 = 0.9$ and $\beta_2 = 0.999$. The learning rate was set to $1 \times 10^{-3}$ and training proceeded for 300 epochs with a batch size of 4 slices. All experiments were implemented in PyTorch 2.0 and conducted on an NVIDIA A100 GPU.

\subsubsection{Data Split}
The 5,600 total slices ($10 \text{ subjects} \times 8 \text{ echoes} \times 70 \text{ slices}$) were split using a subject-based strategy to prevent data leakage. Specifically, 5,040 slices (90\%) from subjects S001-S009 across all echoes were used for training, while 560 slices (10\%) from subject S012 (echoes 1-2) comprised the test set. This ensures our evaluation simulates deployment on unseen patients.

\section{Experiments}

\subsection{Dataset}
We evaluate our method on clinical brain MRI data provided by Washington University Medical School. The dataset contains scans from 10 patients, with 8 echoes per subject and 70 axial slices per echo, yielding 5,600 total slices. Data were acquired using a 64-channel head coil with k-space dimensions of $[64, 234, 176]$ (coils $\times$ height $\times$ width). Both 2x and 4x accelerated acquisitions are available for each subject.

All data underwent standard preprocessing, including intensity normalization to the $[0, 1]$ range computed independently per echo. Complex-valued data were preserved throughout the pipeline, and no additional data augmentation was applied to maintain consistency with clinical acquisition protocols.

\subsection{Evaluation Metrics}
We assess reconstruction quality using PSNR (Peak Signal-to-Noise Ratio) to measure pixel-wise accuracy and SSIM (Structural Similarity Index) to evaluate perceptual quality. Additionally, we perform qualitative visual assessment comparing artifact patterns and image smoothness. The ground truth for all evaluations is the 2x SENSE reconstruction, representing the current clinical standard for accelerated imaging. We compare our method against SENSE reconstruction using pre-computed sensitivity maps from separate calibration, as well as zero-filled reconstruction (inverse Fourier transform of undersampled data) to illustrate the severity of undersampling artifacts.

\subsection{Results}

\begin{figure}[t]
  \centering
  \footnotesize
  
  \begin{tabular}{@{}c@{\hspace{0.5em}}c@{\hspace{0.5em}}c@{\hspace{0.5em}}c@{\hspace{0.5em}}c@{}}
    & Real 4x & Model Output & SENSE & GT \\[0.2em]
    
    \raisebox{-0.5\height}{\rotatebox{90}{Echo 1 Slice 110}} &
    \begin{tabular}{@{}c@{}}
      \includegraphics[width=0.18\textwidth]{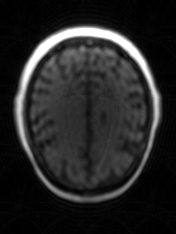} \\
      \footnotesize 29.30 / 0.677
    \end{tabular} &
    \begin{tabular}{@{}c@{}}
      \includegraphics[width=0.18\textwidth]{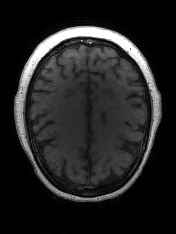} \\
      \footnotesize 36.25 / 0.925
    \end{tabular} &
    \begin{tabular}{@{}c@{}}
      \includegraphics[width=0.18\textwidth]{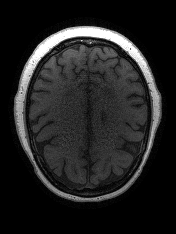} \\
      \footnotesize 37.26 / 0.934
    \end{tabular} &
    \begin{tabular}{@{}c@{}}
      \includegraphics[width=0.18\textwidth]{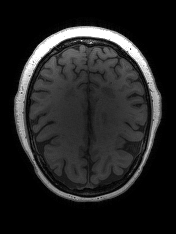} \\
      \footnotesize PSNR/SSIM
    \end{tabular} \\[0.5em]
    
    \raisebox{-0.5\height}{\rotatebox{90}{Echo 1 Slice 120}} &
    \begin{tabular}{@{}c@{}}
      \includegraphics[width=0.18\textwidth]{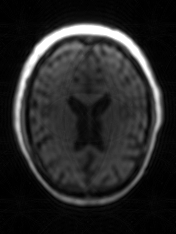} \\
      \footnotesize 28.96 / 0.671
    \end{tabular} &
    \begin{tabular}{@{}c@{}}
      \includegraphics[width=0.18\textwidth]{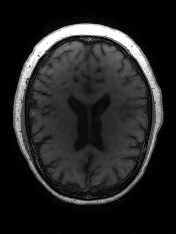} \\
      \footnotesize 35.95 / 0.916
    \end{tabular} &
    \begin{tabular}{@{}c@{}}
      \includegraphics[width=0.18\textwidth]{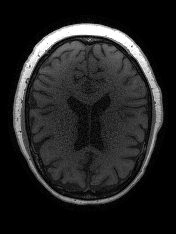} \\
      \footnotesize 37.06 / 0.923
    \end{tabular} &
    \begin{tabular}{@{}c@{}}
      \includegraphics[width=0.18\textwidth]{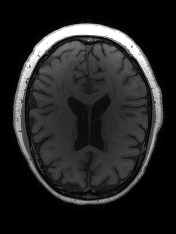} \\
      \footnotesize PSNR/SSIM
    \end{tabular} \\
  \end{tabular}
  
  \caption{Representative reconstructions from Subject S012, Echo 1, Slices 110 and 120. Normalized for visual output. Metrics shown as PSNR (dB) / SSIM.}
  \label{fig:comparison_grid}
\end{figure}

\subsubsection{Quantitative Performance}
Table~\ref{tab:metric_eval} presents quantitative results averaged over the test set comprising 140 slices from subject S012 (echoes 1-2). Our method trained on simulated 4x data achieves performance close to SENSE (within 0.9 dB PSNR), demonstrating that deep learning can match traditional methods when training and testing conditions align. However, performance degrades significantly when applying the model to real 4x acquisitions, with PSNR dropping by 1.8 dB and SSIM by 0.021 compared to simulated data.

\subsubsection{Qualitative Analysis}
Figure~\ref{fig:comparison_grid} shows representative reconstructions from two slices (Echo 1, Slices 110 and 120 from Subject S012). Visual inspection reveals several important characteristics. SENSE output exhibits a speckled noise pattern throughout the image, particularly visible in uniform tissue regions. While this preserves high-frequency information (contributing to higher PSNR/SSIM scores), the speckle may be visually distracting for clinical interpretation. In contrast, our method produces smoother reconstructions with reduced speckle, though with slightly blurred fine details. Despite lower PSNR/SSIM scores, the smoother appearance may be preferable for certain clinical applications where speckle noise is visually distracting. The real 4x input shows the expected undersampling artifacts with aliasing and reduced image quality, which our model successfully mitigates in the reconstruction.

\subsubsection{Simulated vs. Real 4x Data Analysis}
The performance gap between simulated and real 4x data (1.8 dB PSNR, 0.021 SSIM) highlights a critical challenge in deploying learning-based reconstruction. We investigated potential causes and identified three primary factors. First, spatial misalignment occurs because real 2x and 4x acquisitions were collected in sequential scanning sessions, introducing patient motion and scanner drift; even sub-pixel shifts significantly impact training when using pixel-wise loss functions. Second, k-space distribution shifts arise from several sources: different ACS region sizes ($32 \times 32$ for 2x vs. $48 \times 48$ for 4x), physiological changes between scan sessions, and subtle scanner parameter variations. Third, despite applying per-echo normalization to account for intensity variations, residual distribution mismatches persist between the simulated and real accelerated acquisitions.

\section{Discussion}

\subsection{Key Findings}
Our proposed framework successfully demonstrates that joint coil sensitivity estimation and image reconstruction can be performed end-to-end from undersampled k-space data alone. The two-module design enables independent optimization of each component, providing flexibility for future architectural improvements. When training and testing conditions are well-matched (simulated 4x data), our method achieves performance competitive with SENSE while producing visually smoother results.

\subsection{Visual Quality vs. Quantitative Metrics}
An interesting finding is the disconnect between quantitative metrics (PSNR/SSIM) and visual quality. Our method consistently produces smoother reconstructions with less speckle noise compared to SENSE, yet scores lower on both metrics. This suggests that PSNR and SSIM may not fully capture perceptual quality preferences, particularly regarding noise texture. The speckled appearance of SENSE output, while preserving high-frequency information that improves SSIM scores, may be less desirable for radiological interpretation. Future work should incorporate perceptual loss functions or consider alternative evaluation metrics more aligned with clinical preferences.

\subsection{The Sim-to-Real Gap}
The performance degradation on real 4x data represents a key limitation of our current approach. This gap stems from fundamental mismatches between training (2x-based ground truth) and deployment (real 4x input) conditions. Temporal misalignment is the first major factor: 2x and 4x acquisitions were collected sequentially rather than simultaneously, introducing patient motion and physiological changes. K-space distribution shifts also contribute significantly, arising from different ACS region sizes ($32 \times 32$ for 2x vs. $48 \times 48$ for 4x) and scanner reconstruction pipelines that create systematic distribution differences. Finally, despite per-echo normalization, residual intensity calibration differences persist between acceleration factors.

\subsection{Advantages and Limitations}
Despite challenges with real data, our framework offers several practical benefits. The approach requires only undersampled k-space as input, eliminating the need for separate sensitivity calibration scans that increase total acquisition time. The modular architecture design allows CSM estimation and reconstruction modules to be upgraded independently, providing flexibility for incorporating future improvements. The framework can accommodate different acceleration factors by retraining with appropriate undersampling masks.

Current limitations include our reliance on 2x SENSE as ground truth, which may itself contain artifacts that propagate through training. The evaluation is limited to brain MRI anatomy, and generalization to other anatomical regions remains untested. The sim-to-real performance gap represents the most significant challenge for clinical deployment.

\subsection{Future Directions}
Several promising research directions could address current limitations and improve performance. Registration-based alignment techniques could substantially reduce the sim-to-real gap by implementing deformable registration between 2x and 4x acquisitions before training, correcting for patient motion and scanner drift. Domain adaptation approaches, such as adversarial training or cycle-consistency losses, could help the model adapt to distribution shifts between simulated and real accelerated data without requiring perfectly aligned ground truth.

Scaling to larger models with 50-100M parameters and appropriate regularization may improve reconstruction quality for complex anatomical features, though this must be balanced against computational constraints. Multi-task learning across multiple acceleration factors (2x, 3x, 4x) could improve robustness and generalization across different undersampling patterns. Advanced architectures like deep unfolding networks or plug-and-play frameworks that incorporate domain knowledge while maintaining end-to-end training could provide better theoretical guarantees and improved performance.

Incorporating perceptual loss functions, such as VGG-based perceptual losses or adversarial training, could better align optimization with visual quality preferences and potentially resolve the disconnect between quantitative metrics and visual quality. Most critically, clinical validation through radiologist-based assessment of diagnostic quality is essential for deployment, as PSNR and SSIM may not accurately reflect diagnostic value in clinical practice.

\section{Conclusion}
We present an end-to-end deep learning framework for parallel MRI reconstruction from 4x accelerated acquisitions. Our two-module architecture jointly estimates coil sensitivity maps and reconstructs images using only undersampled k-space measurements, eliminating the need for separate sensitivity calibration. On simulated 4x data with perfect alignment to ground truth, our method achieves performance competitive with SENSE (within 0.9 dB PSNR) while producing visually smoother reconstructions. 

We identified important challenges in deploying learning-based reconstruction, particularly the performance gap between simulated and real accelerated data stemming from spatial misalignment and k-space distribution shifts. This work highlights both the promise and practical challenges of learning-based parallel MRI reconstruction. While deep learning can match or exceed traditional methods under ideal conditions, practical deployment requires careful attention to data acquisition protocols, alignment procedures, and domain shift mitigation strategies.

The modular nature of our framework provides a foundation for incorporating advanced reconstruction techniques such as deep unfolding and plug-and-play architectures. Future work will focus on registration-based alignment of multi-acceleration data, domain adaptation techniques, and exploration of perceptual loss functions to bridge the sim-to-real gap and better align quantitative metrics with clinical visual quality preferences. With continued development addressing these challenges, learning-based reconstruction has significant potential to accelerate clinical MRI while maintaining or improving diagnostic image quality.

\begin{ack}
This work was conducted during the Washington University Summer Engineering Fellowship (WUSEF). I thank Dr. Ulugbek S. Kamilov, Shirin Shoushtari, and the Computational Imaging Group for their guidance, and Dr. Dmitriy A. Yablonskiy and the Washington University Medical School Radiology Department for providing the clinical MRI data.
\end{ack}


\bibliographystyle{plain}


\end{document}